# LARGE LANGUAGE MODELS AND (NON-)LINGUISTIC RECURSION

MAKSYMILIAN DĄBKOWSKI, GAŠPER BEGUŠ
UNIVERSITY OF CALIFORNIA, BERKELEY
MANUSCRIPT *as of* JUNE 12, 2023

**ABSTRACT**  Recursion is one of the hallmarks of human language. While many design features of language have been shown to exist in animal communication systems, recursion has not. Previous research shows that GPT-4 is the first large language model (LLM) to exhibit metalinguistic abilities (Beguš, Dąbkowski, and Rhodes, 2023). Here, we propose several prompt designs aimed at eliciting and analyzing recursive behavior in LLMs, both linguistic and non-linguistic. We demonstrate that when explicitly prompted, GPT-4 can both produce and analyze recursive structures. Thus, we present one of the first studies investigating whether meta-linguistic awareness of recursion—a uniquely human cognitive property—can emerge in transformers with a high number of parameters such as GPT-4.

## 1 INTRODUCTION

Recursion is a process or pattern that repeats in a self-similar or self-referential manner. In linguistics, recursion refers to the embedding of phrases within phrases of the same type. This provides linguistic units with potentially infinite layers of depth and complexity.

Recursion is—to the best of our knowledge—one of the few properties entirely unique to human language (Hockett, 1960). Despite previous claims to the contrary (Fitch and Hauser, 2004; Gentner et al., 2006), no other animal communication system has been convincingly shown to feature recursion (Beecher, 2021; Corballis, 2007). For this reason, recursion has become one of the most widely studied, but also hotly debated aspects of language (Chomsky, 2014; Hauser, Chomsky, and Fitch, 2002 vs. Everett, 2005; Jackendoff and Pinker, 2005).

At the same time, recent advances in the development of large language models (henceforth LLMs)—notably OpenAI's (2023) release of GPT-4—have sparked a discussion about the nature of LLMs' generativity and its relation to the human language faculty. Haider (2023) demonstrates that ChatGPT behaves like a native speaker with respect to a number of English and German grammatical competence tasks. Piantadosi (2023) argues that the success of LLMs, capable of remarkable language performance despite a lack of innate biases, refutes Chomsky's (1957) nativist theory of language. Katzir (2023) counters by pointing out that LLMs do not (i) develop human-like linguistic constraints, (ii) derive the attested linguistic typology, (iii) show competence beyond initial performance, (iv) distinguish between likelihood and grammaticality, or (v) reason in a human-like way. Hu and Levy (2023) show that LLMs' probability distributions over strings is a better measure of language performance than prompting (i.e. prompt-based methodologies may underestimate the models' language competence). Beguš, Dąbkowski, and Rhodes (2023) demonstrate that GPT-4 may be the first LLM capable not only of using language, but also being able to perform meta-linguistic tasks by applying linguistic theory to construct valid analyses.

In this paper, we connect the two lines of research by asking whether two recent GPT models (GPT-4 and GPT-3.5) are capable of producing, identifying, and analyzing recursive structures. Previous research focuses on LLMs' implicit capacity for recursion. For example, Lakretz, Desbordes, et al. (2022) and Lakretz, Hupkes, et al. (2021) look at the output probabilities for correct subject-verb agreement in embedded dependencies. They find that the performance of LLMs on long-range dependencies is below chance and conclude that the syntactic processing in transformers is much less robust than in humans. Our methodology is novel in that we ask the models to generate and analyze recursive structures explicitly.

We design prompts aimed at eliciting and analyzing recursion in the domains of language and visual representation. We find that GPT-4 largely succeeds on the tasks in both domains. We compare the performance of GPT-4 with GPT-3.5. We find that the older model succeeds only on the simpler linguistic tasks and more often fails at generating recursive images. We conclude that GPT-4 may be the first LLM capable of general recursive reasoning.

## 2 LINGUISTIC RECURSION

In this section, we test GPT-4's performance on tasks related to linguistic recursion. The tasks involve defining and illustrating recursion, adding more layers of embedding, drawing syntactic trees which capture recursive structure, and writing code that mimics linguistic recursion. We find that GPT-4 performance on the tasks is very good, although the model makes mistakes in syntactic analyses of recursive structures. We compare the performance of GPT-4 and GPT-3.5. We find that while GPT-3.5 can illustrate recursion and has a limited ability to embed clauses, only GPT-4 succeeds on the more elaborate tasks.

First, we prompted the model to provide a definition of recursion and illustrate it (1).[1] GTP-4's answer is correct and fully satisfactory.

> (1) What is recursion in natural language? Define and illustrate.
>
> Recursion in natural language is a property that allows a sentence or a phrase to be expanded indefinitely by embedding the same structure within itself. This linguistic principle is one of the foundations of generative grammar, a theory of language pioneered by Noam Chomsky.
>
> To understand this better, let's consider a simple sentence:
> - "John likes pizza."
>
> Now, let's embed another sentence structure within this sentence:
> - "John, who lives in New York, likes pizza."
>
> We've inserted another sentence or clause ("who lives in New York") within the original sentence. This clause has the same structure as a sentence — it has a subject ("who") and a verb ("lives").
>
> And you can keep embedding more structures indefinitely:
> - "John, who lives in New York, which is a city in the United States, likes pizza."

---

[1] Some of the reported prompts and replies have been edited and simplified for the sake of conciseness and ease of presentation. For full documentation of the chat data presented here, see Dąbkowski and Beguš (2023).

- "John, who lives in New York, which is a city in the United States, that was discovered by the Europeans, likes pizza."
- ...and so on.

We then asked the model to identify and label instances of recursion in eleven sentences (2). GPT-4 performance is often perfect; the model correctly labels all the non-recursive sentences, and correctly classifies each instance of recursion in the data set.

(2) For each of the sentences below, tell me if the sentence contains an instance of recursion. If so, identify it and say what kind of recursion it is. Remember that there are different types of recursion. In addition to clausal recursion, there is also recursive application of adjectival modification, recursive adverbial modification, recursive genitive (possessive) recursion, and so on.

1. Mark said hi.
2. Mark said that Jolene said hi.
3. I saw a big blue elephant.
4. I saw a big elephant.
5. Martha's dog barked.
6. A cat bit Martha's dog's tail.
7. The light flashed quickly.
8. The light flashed incredibly quickly.
9. A boy who had a lollipop yelled my name.
10. A boy yelled my name.
11. A boy who I saw a girl liked yelled my name.

1. "Mark said hi." - This sentence does not contain any form of recursion.
2. "Mark said that Jolene said hi." - This sentence contains clausal recursion, with one clause embedded inside another ("Mark said" containing "that Jolene said hi").
3. "I saw a big blue elephant." - This sentence contains recursive adjectival modification. The adjective "big" modifies "blue elephant", and "blue" further modifies "elephant".
4. "I saw a big elephant." - This sentence does not contain any form of recursion.
5. "Martha's dog barked." - This sentence does not contain any form of recursion.
6. "A cat bit Martha's dog's tail." - This sentence contains recursive genitive (possessive) recursion. "Martha's dog's tail" involves two layers of possessive marking: Martha possessing the dog, and the dog possessing the tail.
7. "The light flashed quickly." - This sentence does not contain any form of recursion.
8. "The light flashed incredibly quickly." - This sentence contains recursive adverbial modification. The adverb "incredibly" modifies another adverb "quickly".

9. "A boy who had a lollipop yelled my name." - This sentence contains clausal recursion, with the relative clause "who had a lollipop" modifying "boy".

10. "A boy yelled my name." - This sentence does not contain any form of recursion.

11. "A boy who I saw a girl liked yelled my name." - This sentence contains clausal recursion. The main clause "A boy ... yelled my name" contains an embedded clause "who I saw a girl liked" which itself contains another embedded clause "a girl liked". The recursion here involves embedding of relative clauses within relative clauses.

Next, we asked GPT-4 to add additional layers of recursion to all the sentences previously identified as recursive (3). The model's performance on all of the examples is perfect.

(3) For each of the sentences you identified as recursive, first add one layer of recursion, and then add another one. Make sure that in adding layers of recursion, you embed structures of the same type. For example, if you identified possessive recursion, add an additional layer of possession. If you identified causal recursion, add a further embedded clause, etc.

2. Original: "Mark said that Jolene said hi."
   +1 layer: "Mark said that Jolene said that John said hi."
   +2 layers: "Mark said that Jolene said that John said that Lisa said hi."

3. Original: "I saw a big blue elephant."
   +1 layer: "I saw a big blue majestic elephant."
   +2 layers: "I saw a big blue majestic beautiful elephant."

6. Original: "A cat bit Martha's dog's tail."
   +1 layer: "A cat bit Martha's neighbor's dog's tail."
   +2 layers: "A cat bit Martha's neighbor's sister's dog's tail."

8. Original: "The light flashed incredibly quickly."
   +1 layer: "The light flashed incredibly almost blindingly quickly."
   +2 layers: "The light flashed incredibly almost blindingly astoundingly quickly."

9. Original: "A boy who had a lollipop yelled my name."
   +1 layer: "A boy who had a lollipop that was cherry-flavored yelled my name."
   +2 layers: "A boy who had a lollipop that was cherry-flavored that his grandmother gave him yelled my name."

11. Original: "A boy who I saw a girl liked yelled my name."
    +1 layer: "A boy who I saw a girl that a dog followed liked yelled my name."
    +2 layers: "A boy who I saw a girl that a dog that a cat chased followed liked yelled my name."

The model correctly adds layers of causal complementation (3.2), adjectival modification (3.3), adverbial modification (3.6), possession (3.8), and clausal modification (3.9,11). We bring special attention to the last example (3.11), where GPT-4 constructs a triply center-embedded sentence. Center embedding, while grammatical, is difficult to process. As such, center-embedding sentences are exceedingly rare and were most likely absent from (or only sparsely present in) the model's training data. GPT-4's ability to construct center-embedded sentences without being explicitly prompted to do so thus suggests that the model acquired grammatical structure beyond simple distributional tendencies of its training data set.

To test the consistency of the model's responses, we repeated the same prompt multiple times. GPT-4 makes occasional errors. However, when prompted to evaluate its own reply, the model often identifies the problems and fixes them. On one occasion, GPT-4 attempts to increase the depth of possessive embedding by adding an adjective instead of a Saxon genitive (4). When prompted to evaluate its answer, the model recognizes it mistakenly changed the type of recursion (instead of adding depth to the existing type) and corrects its response (5).

(4)  For each of the sentences you identified as recursive, first add one layer of recursion, and …

   6. A cat bit Martha's dog's tail.
      - One layer added: "A cat bit Martha's dog's fluffy tail."
      - Two layers added: "A cat bit Martha's dog's fluffy tail's tip."

(5)  Are there any mistakes in your previous response? If so, correct them.

My apologies for the confusion. In an effort to maintain the same kind of recursion, some of my examples changed type of recursion rather than added depth to the existing type. Let's correct these:

   6. A cat bit Martha's dog's tail.
      - One layer added: "A cat bit Martha's neighbor's dog's tail."
      - Two layers added: "A cat bit Martha's neighbor's brother's dog's tail."

We then asked GPT-4 to draw trees for some of the simpler recursive structures it generated using the `forest` package (Živanović, 2017) in LaTeX (6). The analyses provided by the model are generally correct, although there are minor mistakes. For example, the model represents proper nouns and pronouns and unary branching NP → N structures (as opposed to treating them as lexical NPs). In (6b), GPT-4 labels each projection of adjectival adjunction as AP (instead of the correct N'). Despite these errors, however, the model correctly represents the hierarchical recursive structure of the sentences at hand.

(6)  Using the `forest` package in LaTeX, draw a syntax tree for the sentence:
   a. "Mark said that Jolene said that John said that Lisa said hi."
   b. "I saw a big blue majestic beautiful elephant."

Assume X-bar theory.

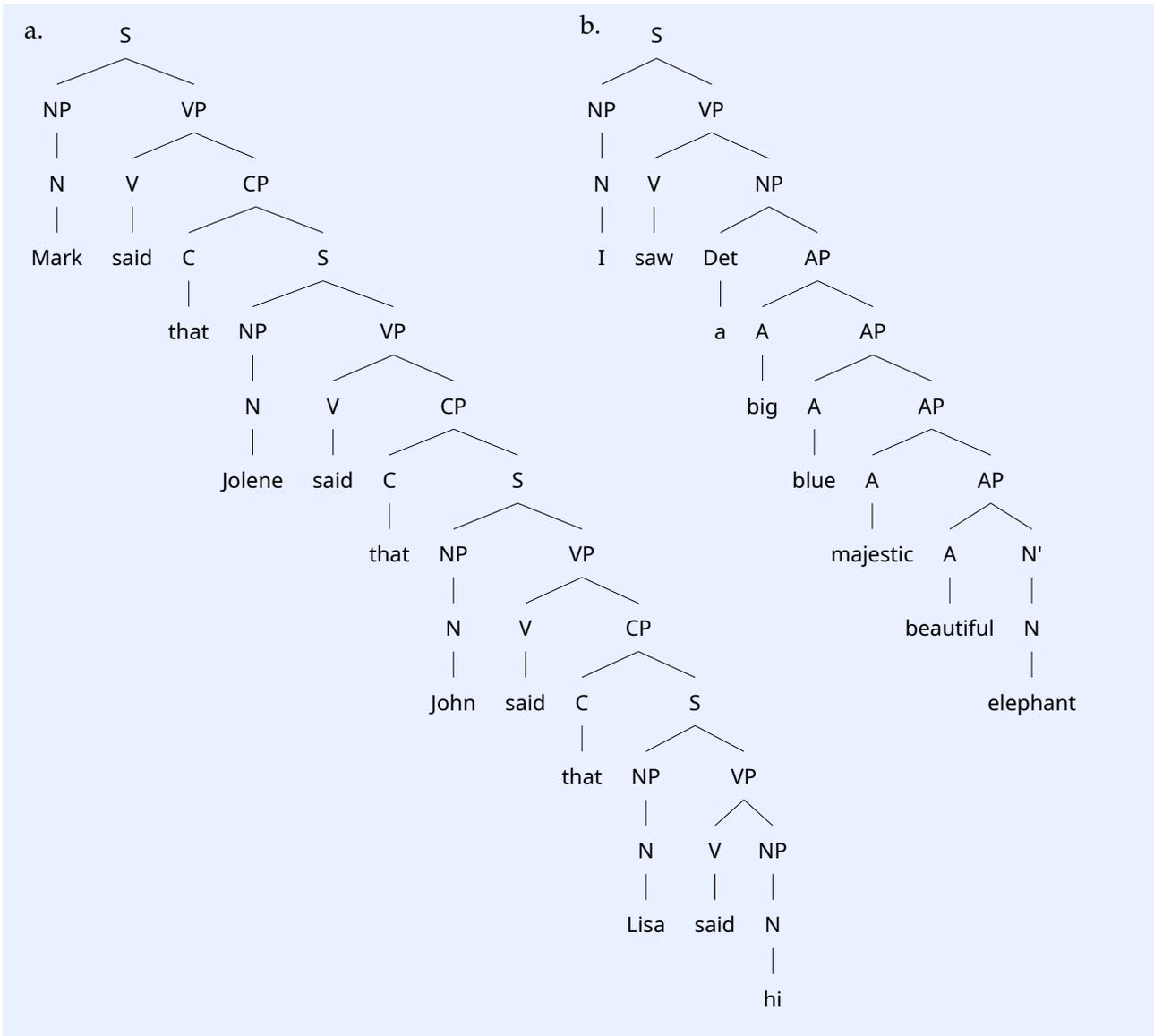

GPT-4 has a harder time drawing correct syntactic trees for some of the other sentences. For example, when prompted to represent possessive recursion, the model generates an incorrect flat structure (7a). However, the model is capable of refining its answers. Simply asking it if the tree is correct results in a remarkable improvement; GPT-4 redraws the tree with the complex possessor represented as a left-branching structure (7b). The tree still leaves much to be desired— the possessive clitic *'s* is represented as a node dominating the possessor and the input chunk *dog's tail* has been turned into a compound noun *dog tail*. At last, prompting the model to evaluate the second tree results in another round of improvement, with each level or recursive possession represented as left-branching and binary (7c).

(7)  a. Using the `forest` package in LaTeX, draw a syntax tree for the sentence "A cat bit Martha's neighbor's sister's dog's tail." Assume X-bar theory.
  b. Is the tree you drew correct?

c. Is the tree you drew correct?

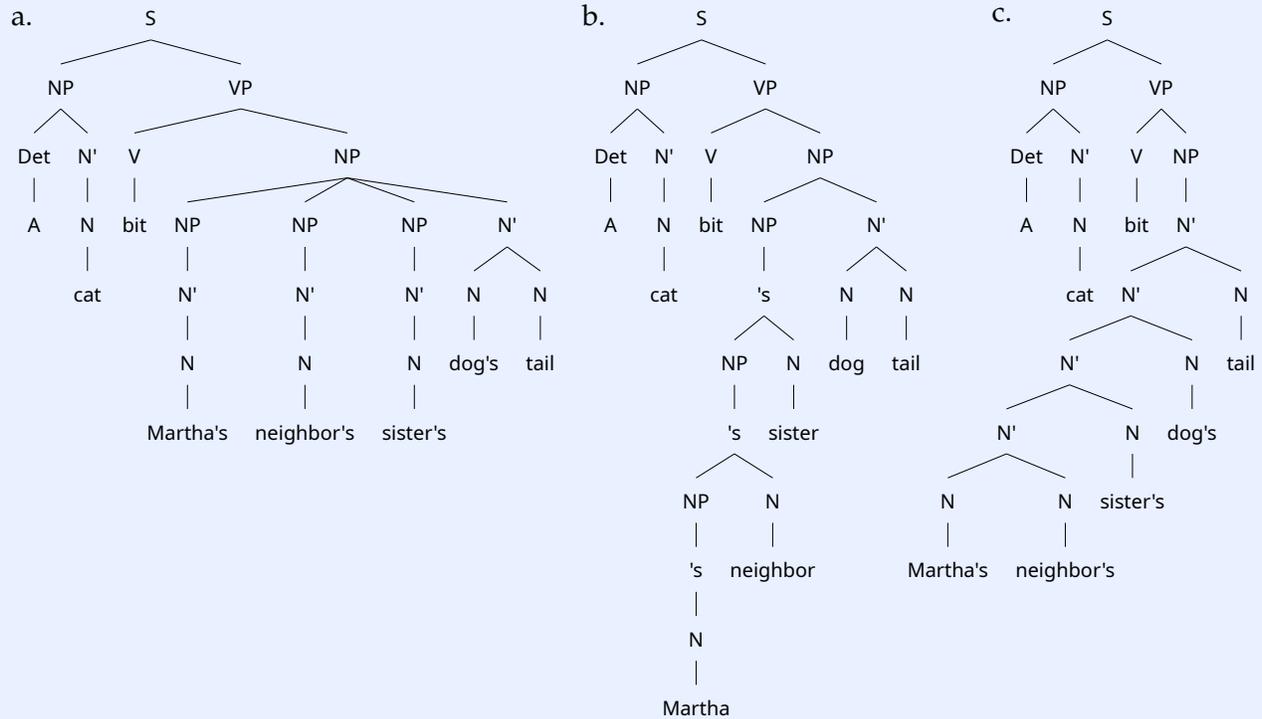

In the next task, we tested GPT-4's ability to model linguistic recursion with code. First, we asked the model to volunteer a piece of recursive code (8). The model offered a simple recursive function in Python calculating the factorial of a number.

(8)  Write a piece of recursive code.

```python
def factorial(n):
    # Base case: factorial of 0 or 1 is 1
    if n == 0 or n == 1:
        return 1
    # Recursive case: n * factorial of n-1
    else:
        return n * factorial(n-1)

print(factorial(5))  # Output: 120
```

Finally, we prompted the model to write a piece of code isomorphic to how recursion works in natural languages. GPT-4 succeeded on the task by writing a function that models English recursive speech report (9).

(9) Write a piece of recursive code that is isomorphic to how recursion works in natural languages.

In natural languages, recursion is often represented in nested structures like subordinate clauses. For instance, consider this sentence:

"Alice said that Bob said that Charlie said that he didn't take the cake."

The structure of this sentence can be represented recursively as "Person said that [sentence]". This pattern can continue to nest indefinitely.

Here's a Python function that simulates this kind of recursion:

```python
def recursive_sentence(names, message, depth=0):
    # Base case: if depth == 0, return the message
    if depth == 0:
        return message
    # Recursive case: "{name} said that {sentence}"
    else:
        return f"{names[depth % len(names)]} said that {recursive_sentence(names,
            message, depth - 1)}"

names = ["Alice", "Bob", "Charlie"]
message = "he didn't take the cake"

print(recursive_sentence(names, message, 3))  # Output: "Alice said that Bob said
    that Charlie said that he didn't take the cake"
```

In interim summary, GPT-4 is capable of identifying, producing, and analyzing recursive structures. While GPT-4's performance is not always perfect, the model can often auto-correct when prompted to evaluate its own output.

Next, we compared the performance of GPT-4 with an older version of the model, GPT-3.5. GPT-3.5 performs well on the basic tasks of defining and illustrating recursion (10-11).

(10) What is recursion in natural language? Define and illustrate.

Recursion in natural language refers to the linguistic phenomenon where a sentence or phrase contains a repeated structure or pattern within itself. It involves the repetition of a particular linguistic element or construction within a larger linguistic context. In simpler terms, recursion is a form of self-embedding, where a linguistic unit is embedded within another unit of the same kind.

To illustrate recursion, let's consider an example sentence: "The cat that caught the mouse that scared the dog that chased the squirrel that climbed the tree..."

In this sentence, the relative clause "that caught the mouse" is embedded within the main clause "The cat," creating a recursive structure. Furthermore, the relative clause itself contains another embedded clause "that scared the dog," which is embedded within the previous relative clause. This pattern of embedding continues as each new clause is introduced.

(11) Write a complete recursive sentence with multiple levels of recursion.

"The book that the librarian recommended to the student who borrowed it from the library that the author praised for its comprehensive content that inspired the professor to write a research paper that the journal published and gained recognition for its insightful analysis that influenced the readers to explore new perspectives on the subject."

However, GPT-3.5 is markedly inferior to the latest model when it comes to identifying and labeling instances of recursion. Given the same (below elided) prompt as GPT-4 in (2), GPT-3.5 fails on many of the same example sentences that GPT-4 succeeded at (12).

(12) For each of the sentences below, tell me if the sentence contains an instance of recursion. If ...

1. This sentence does not contain recursion.
2. This sentence contains clausal recursion. The main clause "Mark said" embeds the subordinate clause "that Jolene said hi."
3. This sentence does not contain recursion.
4. This sentence does not contain recursion.
5. This sentence does not contain recursion.
6. This sentence contains recursive genitive (possessive) recursion. The phrase "Martha's dog's tail" involves the possessive form "Martha's" applied twice to indicate the possession relationship between "Martha," "dog," and "tail."
7. This sentence does not contain recursion.
8. This sentence does not contain recursion.
9. This sentence contains recursive adjectival modification. The adjective "big" modifies the noun "elephant," and the adjectival phrase "who had a lollipop" further modifies the noun "boy."
10. This sentence does not contain recursion.
11. This sentence contains clausal recursion. The main clause "A boy liked" embeds the subordinate clause "who I saw a girl."

The model correctly identifies clausal recursion in (12.2) and (12.11), as well as lack of recursion in (12.1), (12.4), (12.5), (12.7), and (12.10). However, GPT-3.5 fails to note adjectival recursion in (12.3) and adverbial recursion in (12.8). The model recognizes possessive recursion in (12.6), but gives a garbled explanation of the recursive structure, saying that "the possessive form *Martha's*" is "applied twice to indicate the possession relationship between *Martha*, *dog*, and *tail*." GPT-3.5 observes recursion in (12.9), but calls it "recursive adjectival modification." The model correctly notes that the "phrase *who had a lollipop* ... modifies the noun *boy*," but also splices in a comment about a previous sentence, saying that the "adjective *big* modifies the noun *elephant*."

We then asked GPT-3.5 to add additional layers of recursion to all the sentences previously identified as recursive (13). Here again, we see that GPT-3.5 is outperformed by GPT-4.

(13) For each of the sentences you identified as recursive, first add one layer of recursion, and then add another one. Make sure that in adding layers of recursion, you embed structures of the same type. For example, if you identified possessive recursion, add an additional layer of possession. If you identified causal recursion, add a further embedded clause, etc.

2. Mark said that Jolene said that Mark said hi.
   This sentence contains clausal recursion. The main clause "Mark said" embeds the subordinate clause "that Jolene said," and within that subordinate clause, there is another embedded clause "that Mark said hi."
6. A cat bit Martha's dog's tail's end.
   This sentence contains recursive genitive (possessive) recursion. The phrase "Martha's dog's tail" involves the possessive form "Martha's" applied twice. By adding another layer, we have "tail's end," where the possessive form "tail's" is applied to indicate the end of the tail.
9. A boy who had a lollipop that a girl who had a hat gave yelled my name.
   This sentence contains recursive adjectval modification and clausal recursion. The adjective phrase "who had a lollipop" modifies the noun "boy," and within that phrase, there is another embedded clause "that a girl who had a hat gave."

First, despite having identified four recursive sentences (12.2, 6, 9, 11), GPT-3.5 only attempts the task on the first three of them. Second, the model ignores the request to "first add one layer of recursion, and then add another one;" instead, the model only adds one layer of recursion. GPT-3.5 correctly adds further embedding in (13.2) and (13.6). However, in (13.6), the model still claims that the "phrase *Martha's dog's tail* involves the possessive form *Martha's* applied twice." In (13.9), GPT-3.5 attempts to embed another relative clause but fails to do. The model generates the ungrammatical "*A boy who had a lollipop that a girl who had a hat gave yelled my name," where the ditransitive verb *give* is used monotransitively. (To make the sentence grammatical, one could, for example, use a pronoun, i.e. "A boy who had a lollipop that a girl who had a hat gave *him* yelled my name.")

Next, we tested GPT-3.5's performance on generating tree diagrams which represent syntactic structures. The model succeeds on simpler structures, such as those discussed in (6). (GPT-3.5's outputs are not presented here for reasons of brevity.) However, the model fails at left-branching possessive recursion. When first prompted to draw a syntax tree for the sentence "A cat bit Martha's neighbor's sister's dog's tail," GPT-3.5 generates a right-branching structure and fails to identify the terminal nodes in the recursive NP (14a). (Also observe other errors. For example, the top node is labeled as TP, branching into a single DP. The verb is analyzed as rising to T and no V node is given. Instead, the unary-branching VP dominates the object DP.) Prompting the model to evaluate its previous response does not result in improved performance. When asked whether the tree is correct, GPT-3.5 generates another right-branching structure (14b). When asked to evaluate the tree again, the model generates a baroque confused mess (14c).

(14) a. Using the `forest` package in LaTeX, provide code for a syntax tree for the sentence: "A cat bit Martha's neighbor's sister's dog's tail." Assume X-bar theory.
  b. Is the tree you drew correct?
  c. Is the tree you drew correct?

a.
```
          TP
          |
          DP
         / \
       DP   T'
      /\   /\
     D NP T  VP
     |  |  |  |
     A cat bit DP
              /\
             D  NP
             |  |
             N  NP
                /\
               D  NP
               |  /\
               N D  NP
                 |  /\
                 N D  NP
                   |  |
                   N tail
```

b.
```
          TP
          |
          DP
         / \
       DP   T'
      /\   /\
     D NP T  VP
     |  |  |  |
     A cat bit DP
              |
              DP
             /\
            D  NP
            |  /\
         Martha's N'
                 /\
                N  NP
                |  |
            neighbor's N'
                       /\
                      N  NP
                      |  |
                   sister's N'
                            /\
                           N  NP
                           |  |
                        dog's tail
```

c.
```
                  TP
                 /  \
               DP    T'
              /\    /\
             D' T   VP
            /\  |   |
           D NP bit DP
           |  |    |
           A cat   D'
                   /\
                  D  NP
                  |  |
                  N  DP
                     |
                     D'
                    /\
                   D   NP
                   |   /\
                   N DP  N'
                     |   /\
                     D' N  NP
                    /\ |   /\
                  D NP dog's DP  N'
                  |  |       |  /\
                  N  D'      D' N NP
                     /\     /\ |  |
                    D NP   D NP neighbor's DP   N'
                    | |    | |             |   /\
                    N tail N sister's      D'  N NP
                                          /\   |  |
                                         D NP  A cat
                                         |  |
                                         N Martha's
```

Finally, we tested GPT-3.5's performance on capturing language structure with recursive code. When prompted to generate recursive code, GPT-3.5 implements a factorial function similar to GPT-4's (again, not reported here for brevity). However, when prompted to write a piece of code that is isomorphic natural language recursion (15), the model fumbles and generates a function that takes "Hello, how are you?" as input and returns "Hello, how are you?"

(15)   Write a piece of recursive code that is isomorphic to how recursion works in natural languages.

```python
def speak_recursively(sentence):
    if len(sentence) == 0:
        return ""
    else:
```

```
    first_word = sentence[0]
    rest_of_sentence = sentence[1:]
    return first_word + " " + speak_recursively(rest_of_sentence)
```

Here's an example usage of the `speak_recursively` function:

```
sentence = "Hello, how are you?"
result = speak_recursively(sentence)
print(result)
```

Output:

```
Hello, how are you?
```

In summary, we tested GPT-4 and GPT-3.5 on a number of language-related tasks involving explicit command of recursion. GPT-4 succeeds at defining, illustrating, identifying different types of recursion, and generating deeper levels of syntactic embedding. The model is capable of providing theoretical analyses of various syntactic structures and writing code that mimics linguistic recursion. While GPT-4 occasionally falters, the model is capable of self-correction—oftentimes simply asking it to evaluate its previous answer results in a dramatic improvement.

GPT-3.5 succeeds at defining and illustrating linguistic recursion. However, the model struggles with identifying different types of recursion and generating further layers of recursive structure. GPT-3.5 only seems capable of providing syntactic analyses of the simpler structures. The model does not capture linguistic recursion with computer code. When GPT-3.5 fails at a task, prompting the model to evaluate its output seldom improves performance, and often has the opposite effect.

## 3 VISUAL RECURSION

In this section, we test GPT-4 and GPT-3.5's performance on a number of tasks related to recursion in the visual domain by prompting the models to generate recursive images and diagrams. A recursive image is an image that contains the entire image as a part of itself. We find that GPT-4 succeeds on generating consistently recursive art. The outputs of GPT-3.5 appear more creative regarding the range of generated structures, more often fail to exhibit recursion.

First, we prompted GPT-4 to produce recursive ASCII art. We repeated each prompt several times to observe the diversity of structures generated by each model. When asked for a recursive ASCII diagram, the model outputs variations on a binary-branching tree (16).

We then prompted GPT-4 to draw recursive ASCII images (17). The model often responds with variations on the Sierpiński triangle (17a-b). Other structures include a rectangle with smaller rectangles protruding rightwards (17c), a Christmas-like tree (17d), a rhomboidal fractal, and a wave of increasing amplitude (17e).

(16) Draw a recursive ASCII diagram.

a. 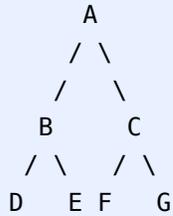

b. 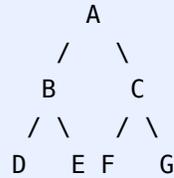

c. 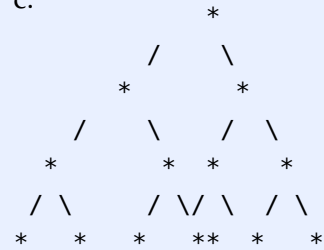

(17) Draw a recursive ASCII image.

a. 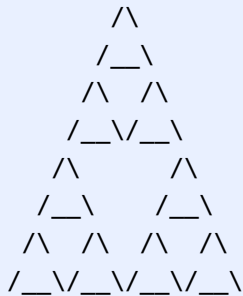

b. 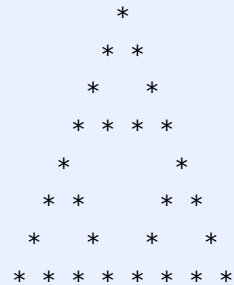

c. 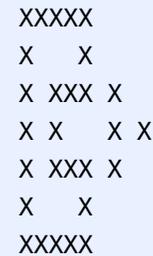

d. 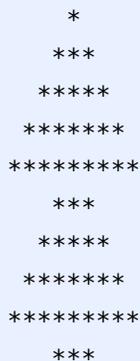

e. 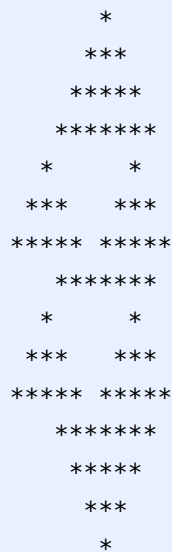

f. 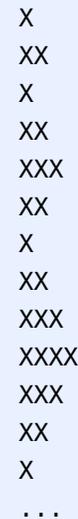

GPT-4's performance on the task is good; most of the images—perhaps with the exception of (17d)—are clearly recursive. While Sierpiński triangles are almost certainly present in the training and therefore likely memorized, the other structures generated by the model are possibly novel.[2]

---

2 We note that although GPT-4 is able to generate recursive images rather successfully, it does not describe them well. For example, the model says that the image in (17c) "shows a larger X shape, with a smaller X shape nested inside it. The Next, we tested the model's performance with the charting and diagramming plugin Show Me. The plugin works by taking the user's request for a visual representation of a concept or process and generating a diagram using Mermaid syntax. To test the plugin's functionality, we repeatedly asked the model for recursive diagrams (18). GPT-4 succeeds on the task and generates a wide variety thereof. The model's outputs include a simple loop (18a), a basic recursive function schema (18b), a simple simple representation of factorization (18c), a more complex representation of factorization (18d), and a binary-branching tree (18e).

(18) Generate a recursive diagram using the Show Me plugin. The diagram can be of anything you'd like. However, you have to make sure the diagram itself is recursive.

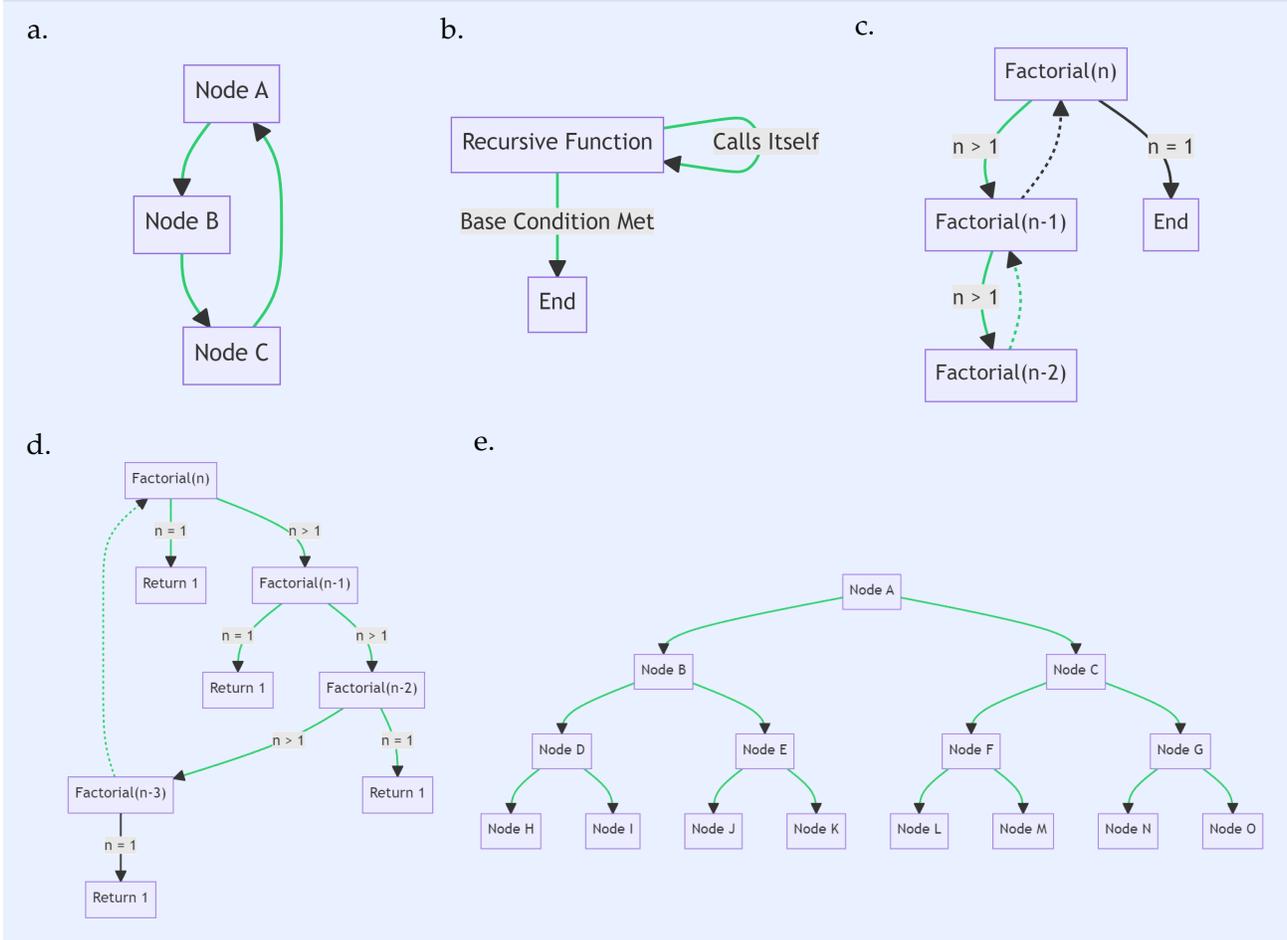

Our last task for GPT-4 was to draw a recursive image using the tikz package (Tantau, 2007) in LaTeX. The model struggles with the task and requires multiple rounds of reprompting. Oftentimes the images generated by the model's code are not recursive, are trivially recursive, or do not compile. Some replies, however, are remarkably successful. In (19), GPT-4 outputs tikz code which contains

---

smaller X is one-third the size of the larger one, following a simple recursive pattern." GPT-4 considers the pattern in (17f) recursive "because it reproduces the original X in each subsequent line, and if you continue the pattern, it will repeat itself infinitely." Sometimes the model's explanations are more successful. For example, GPT-4 characterized the "smaller diamonds" in as "recursively placed within the larger diamond." For complete responses of both models to each of the discussed queries, see Dąbkowski and Beguš (2023).

the recursive function `fractal` producing an image with seven levels of ever-smaller triangles. The compiled image is given on the right.

(19)   Draw a recursive image using LaTeX's `tikz` package.

```
\begin{tikzpicture}
\draw[very thick] (0,0) -- (0,2);
\newcommand{\Y}[2]{
  \draw[#1] (0,0) -- (45:#2) -- (-45:#2) -- cycle;
}
\newcommand{\fractal}[2]{
  \Y{scale=#2}{#2}
  \ifnum#1>0
    \begin{scope}[shift={(45:#2)},rotate=45]
      \fractal{\the\numexpr#1-1\relax}{#2/2}
    \end{scope}
    \begin{scope}[shift={(-45:#2)},rotate=-45]
      \fractal{\the\numexpr#1-1\relax}{#2/2}
    \end{scope}
  \fi
}
\begin{scope}[yshift=2cm]
\fractal{7}{2}
\end{scope}
\end{tikzpicture}
```

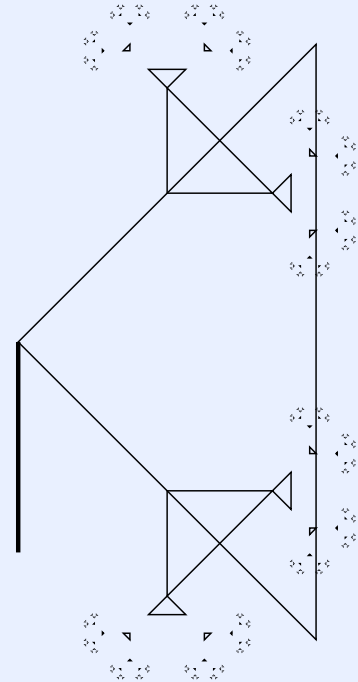

In interim summary, GPT-4 is capable of generating recursive images and diagrams using ASCII, as well as Mermaid and `tikz` syntax. The `tikz` package is the most versatile tool for diagramming and at the same time most challenging for the model.

Finally, we compared the performance of GPT-4 with a previous iteration of the model, GPT-3.5. When asked to draw a recursive ASCII diagram (20), GPT-3.5 comes up with a wider range of structures than GPT-4. In (20a), the model generates a classic Sierpiński triangle. In (20b), the model outputs four connected rectangles. In (20c), the model replies with a honeycomb-like structure. While GPT-3.5's diagrams are generally more varied than GPT-4's, the generated structures are often non-recursive (e. g. 20b).

Next, we prompted GPT-3.5 to draw recursive ASCII images (21). Here again, the model's creations are more diverse and creative than those of GPT-4. While GPT-3.5 often generates Sierpiński triangles (21a), and other triangles (21b), the model also outputs a "house" (a trapezoid on top of a rectangle) within a "house" (21c), a vaguely cloud-like shape (21d), nested rectangles (21e), and two trapezoids within a "house" (a rectangle with yet another trapezoid on top) (21f). Note, however, that although GPT-3.5's outputs are more diverse than GPT-4's, only the (likely memorized) Sierpiński triangle (21a), the house within a house (21c), and the nested rectangles (21e) are clearly compliant with the prompt, which is to say—recursive.[3]

(20) Draw a recursive ASCII diagram.

a. 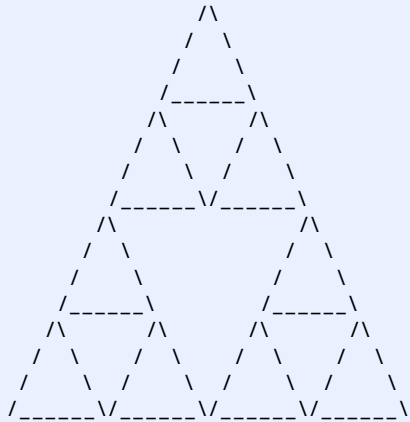

b. 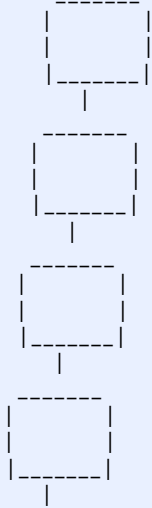

c. 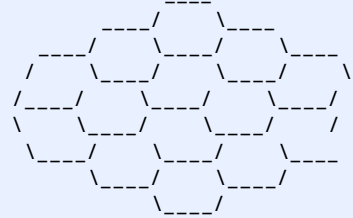

(21) Draw a recursive ASCII image.

a. 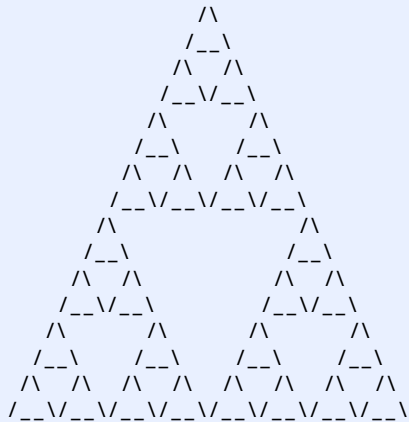

b. 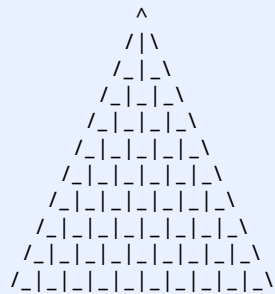

c. 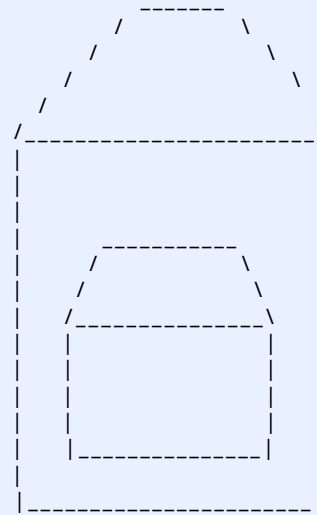

---

[3] Intrigued by the different character of the ASCII art generated by GPT-4 and GPT-3.5, we tested both models' performance with different temperature values. Lowering the temperature values below 1.0 results in more conservative output (and more Sierpiński triangles). Raising the temperature above 1.0 increases the frequency with which elaborate original structures are generated. However, the elaborate structures are less frequently recursive. We conclude that changing the temperature setting does not considerably improve either model's performance.

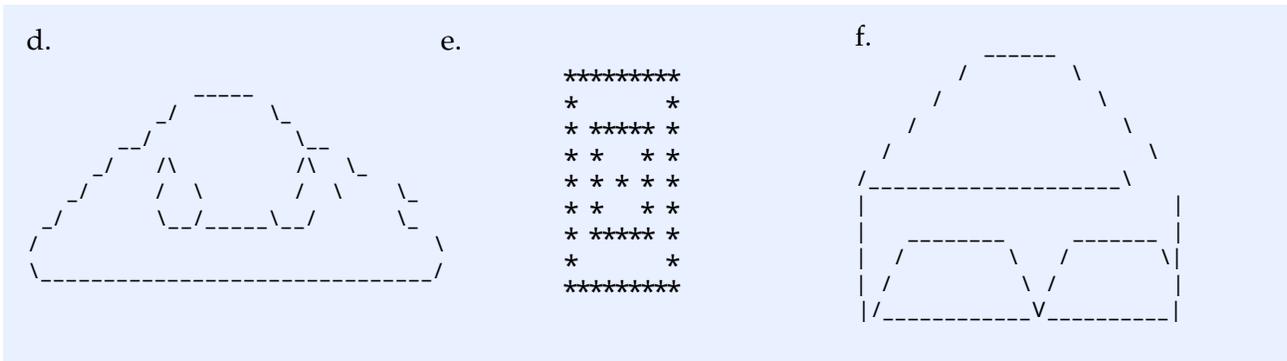

Finally, we prompted GPT-3.5 to draw a recursive image in `tikz`. Similarly to the newer model, GPT-3.5 struggles with `tikz` syntax and most of its outputs do not compile without errors. When successful upon reprompting, the model outputs code which makes use of a `recursive` function to generate an image of a triangle made up of green upward-pointing arrows. Notwithstanding, while the code itself is indisputably recursive, the resulting image is not.

(22)    Draw a recursive image using LaTeX's `tikz` package.

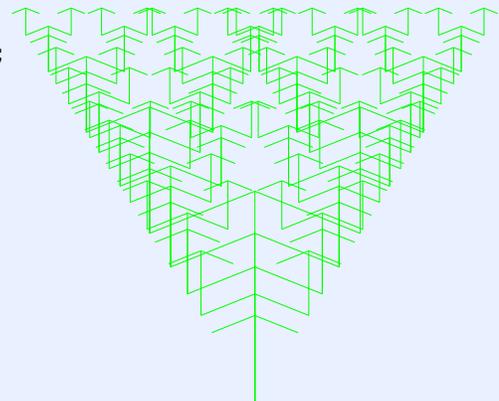

```
\begin{tikzpicture}
% Define the recursive function for drawing the fractal tree
\def\recursive#1#2{
  \ifnum#1>0
    \draw[green] (0,0) -- (0, #2);
    \draw[green] (0, #2) -- (-0.5 * #2, 0.8 * #2);
    \draw[green] (0, #2) -- (0.5 * #2, 0.8 * #2);
    \recursive{\numexpr#1-1\relax}{0.8 * #2}
    \begin{scope}[shift={(-0.5 * #2, 0.8 * #2)}]
      \recursive{\numexpr#1-1\relax}{0.6 * #2}
    \end{scope}
    \begin{scope}[shift={(0.5 * #2, 0.8 * #2)}]
      \recursive{\numexpr#1-1\relax}{0.6 * #2}
    \end{scope}
  \fi
}
% Call the recursive function with initial values
\recursive{5}{4}
\end{tikzpicture}
```

In summary, we tested the performance of GPT-4 and GPT-3.5 on tasks that involved the use of recursion in the visual domain. We find that GPT-4 can generate recursive diagrams and images using ASCII characters, Mermaid syntax (using the Show Me plugin), and `tikz` syntax. GPT-3.5 can also draw diagrams in ASCII and `tikz`, but more frequently fails to generate recursive structures. Both models struggle with the syntax of the `tikz` package and require reprompting to fix their code.

## 4 DISCUSSION AND CONCLUSION

In conclusion, we tested GPT-4 and GPT-3.5's explicit command of recursion on a series of linguistic and visual tasks. We find that GPT-4 is capable of identifying, creating, and analyzing recursive structures in the linguistic domain. The model is also capable of creating recursive art in ASCII, illustrating recursive processes with Mermaid syntax, and creating fractal graphs using the tikz package in LaTeX. GPT-4's success at producing recursion across domains suggests that it may be the first language model with emergent and general (i.e. non-domain-specific) recursive ability. In comparison, GPT-3.5's performance on the same tasks is more limited. While the older model is capable of identifying recursion and illustrating it with linguistic examples, it often fails at analyzing it and representing it visually.

One could object to our findings by noting that the GPT training data almost certainly contains linguistics textbooks (and other pedagogical and research materials) that explicitly discuss recursion. Therefore, the models' understanding of recursion can be attributed to memorization. We respond by observing that humans are not intuitively aware of recursion either and need to be explicitly instructed as well. Thus, there are no fundamental asymmetries in the prior training necessary for humans and LLMs to succeed on the tasks presented in this paper.

Finally, we observe that while GPT-4's initial replies to a prompt are not always correct, the model's performance improves considerably when asked to evaluate its own output (e.g. 5, 7). These results are at odds with Şahin et al.'s (2020) assertion that language models "lack the skill of iterative reasoning upon knowledge" (p. 1241), and challenge Katzir (2023), who shows a case where reprompting does not result in an improvement and argues that "further time and resources are of no use to ChatGPT" (p. 5). Instead, our findings suggest that—given the right experimental set-up—GPT-4 is capable of emergent iterative reasoning, which is a prerequisite for complex problem-solving.

BIBLIOGRAPHY

Beecher, Michael D. (2021). "Why are no animal communication systems simple languages?" In: *Frontiers in Psychology* 12. ISSN: 1664-1078. DOI: 10.3389/fpsyg.2021.602635. URL: https://www.frontiersin.org/articles/10.3389/fpsyg.2021.602635.

Beguš, Gašper, Maksymilian Dąbkowski, and Ryan Rhodes (2023). *Large linguistic models: Analyzing theoretical linguistic abilities of LLMs*. arXiv: 2305.00948 [cs.CL].

Chomsky, Noam (1957). *Syntactic Structures*. The Hague: Mouton and Co.

Chomsky, Noam (2014). "Minimal recursion: Exploring the prospects." In: *Recursion: Complexity in Cognition*. Studies in Theoretical Psycholinguistics (SITP) 43, pp. 1–15.

Corballis, Michael C. (2007). "Recursion, language, and starlings." In: *Cognitive Science* 31.4, pp. 697–704.

Dąbkowski, Maksymilian and Gašper Beguš (June 2023). *Large language models and (non-)linguistic recursion*. OSF. URL: https://osf.io/kjpq8/.

Everett, Daniel L. (2005). "Cultural constraints on grammar and cognition in pirahã: Another look at the design features of human language." In: *Current Anthropology* 46.4, pp. 621–646.

Fitch, W. Tecumseh and Marc D. Hauser (2004). "Computational constraints on syntactic processing in a nonhuman primate." In: *Science* 303.5656, pp. 377–380.

Gentner, Timothy Q., Kimberly M. Fenn, Daniel Margoliash, and Howard C. Nusbaum (2006). "Recursive syntactic pattern learning by songbirds." In: *Nature* 440.7088, pp. 1204–1207.

Haider, Hubert (2023). "Is Chat-GPT a grammatically competent informant?" Manuscript. Salzburg University. URL: https://lingbuzz.net/lingbuzz/007285.

Hauser, Marc D., Noam Chomsky, and W. Tecumseh Fitch (2002). "The faculty of language: What is it, who has it, and how did it evolve?" In: *Science* 298.5598, pp. 1569–1579.

Hockett, Charles F. (1960). "The Origin of Speech." In: *Scientific American* 203.3, pp. 88–97.

Hu, Jennifer and Roger Levy (2023). "Prompt-based methods may underestimate large language models' linguistic generalizations." Manuscript. Massachusetts Institute of Technology. URL: https://lingbuzz.net/lingbuzz/007313.

Jackendoff, Ray and Steven Pinker (2005). "The nature of the language faculty and its implications for evolution of language (Reply to Fitch, Hauser, and Chomsky)." In: *Cognition* 97.2, pp. 211–225.

Katzir, Roni (2023). "Why large language models are poor theories of human linguistic cognition. a reply to Piantadosi (2023)." Manuscript. Tel Aviv University. URL: https://lingbuzz.net/lingbuzz/007190.

Lakretz, Yair, Théo Desbordes, Dieuwke Hupkes, and Stanislas Dehaene (2022). "Can transformers process recursive nested constructions, like humans?" In: *Proceedings of the 29th International Conference on Computational Linguistics*, pp. 3226–3232.

Lakretz, Yair, Dieuwke Hupkes, Alessandra Vergallito, Marco Marelli, Marco Baroni, and Stanislas Dehaene (2021). "Mechanisms for handling nested dependencies in neural-network language models and humans." In: *Cognition* 213, p. 104699.

OpenAI (2023). *Gpt-4 Technical Report*. arXiv: 2303.08774 [cs.CL].

Piantadosi, Steven (2023). "Modern language models refute Chomsky's approach to language." Manuscript. University of California, Berkeley. URL: https://lingbuzz.net/lingbuzz/007180.

Şahin, Gözde Gül, Yova Kementchedjhieva, Phillip Rust, and Iryna Gurevych (2020). "PuzzLing Machines: A challenge on learning from small data." In: *Proceedings of the 58th Annual Meeting of the Association for Computational Linguistics*. Association for Computational Linguistics, pp. 1241–1254. URL: https://aclanthology.org/2020.acl-main.115.pdf.

Tantau, Till (2007). *TikZ and pgf: Manual for version 1.18*. URL: https://www.bu.edu/math/files/2013/08/tikzpgfmanual.pdf.

Živanović, Sašo (2017). *Forest*. https://ctan.org/pkg/forest. Version 2.1.5. LaTeX package for drawing (linguistic) trees.